\documentclass[journal]{IEEEtran}
\usepackage{amsmath,amsfonts,amssymb,bm,mathtools}
\usepackage{algorithmic}
\usepackage[ruled,vlined]{algorithm2e}
\usepackage{array}
\usepackage{booktabs,multirow,makecell,tabularx}
\usepackage{textcomp}
\usepackage{stfloats}
\usepackage{url}
\usepackage{verbatim}
\usepackage{graphicx}
\usepackage{cite}
\usepackage{xcolor}
\usepackage{xspace}
\usepackage{tikz}
\usepackage{stackengine}
\usepackage{etoolbox,xpatch,xstring}
\usepackage{enumerate}
\usepackage{footmisc}
\usepackage[hidelinks]{hyperref}
\usepackage{balance}
\def\BibTeX{{\rm B\kern-.05em{\sc i\kern-.025em b}\kern-.08em
    T\kern-.1667em\lower.7ex\hbox{E}\kern-.125emX}}

\newcommand{\algo}{PGDG}

\definecolor{softsage}{RGB}{150,170,150}

\newcommand{\safeincludegraphics}[2][]{%
  \IfFileExists{#2}{\includegraphics[#1]{#2}}{%
    \fbox{\parbox[c][2.1in][c]{0.97\linewidth}{\centering Missing figure\\\texttt{\detokenize{#2}}}}%
  }%
}
\newcommand{\equalcontrib}{\textsuperscript{\dag}}
\newcommand{\cmu}{\textsuperscript{1}}
\newcommand{\dexmate}{\textsuperscript{2}}

\begin{document}

\title{Physically Grounded Data Generation for Robust Bimanual Policy Learning from a Single Demonstration}

\author{
Cunxi Dai\cmu\equalcontrib,~
Haoran Chang\dexmate\equalcontrib,~
Aditya Nisal\dexmate,~
Rahul Kumar\dexmate,~
Guofei Chen\dexmate,~
Tao Chen\dexmate,~
Yuzhe Qin\dexmate,~
Guanya Shi\cmu%
% \thanks{\textsuperscript{\dagger}Equal contribution.}
\thanks{$^\dagger$ Equal contribution.}
\thanks{This work was partially conducted during an internship at Dexmate.}
\thanks{This work is partially supported by NSF Award \#2512805.}
\thanks{\textsuperscript{1}Cunxi Dai and Guanya Shi are with the Robotics Institute, Carnegie Mellon University, Pittsburgh, PA 15213, USA.
Email: {\tt\small \{cunxid, guanyas\}@andrew.cmu.edu}.}
\thanks{\textsuperscript{2}Haoran Chang, Aditya Nisal, Rahul Kumar, Guofei Chen, Tao Chen, and Yuzhe Qin are with Dexmate, USA.
Email: {\tt\small \{hr.chang, a.nisal, r.kumar, 
gf.chen, tao.chen, yz.qin\}@dexmate.ai}.}
}
\maketitle

\begin{abstract}
Behavior cloning for contact-rich bimanual manipulation remains challenging because diverse demonstrations are expensive to collect, and even small disturbances can push the system into off-manifold states where no recovery supervision is available. We propose PGDG, a data generation framework with zero-shot curation that expands a single demonstration into a compact dataset of physically plausible, successful, and diverse recovery behaviors without additional human labeling. PGDG iterates between a physics-grounded sampler and a dataset curator, where the curator selects informative, non-redundant, and recoverable behaviors to update the sampling distribution toward under-covered recovery modes, and the sampler draws physically plausible rollout candidates from this updated distribution and retains successful trajectories. To further improve data quality, PGDG applies short-horizon sampling-based control to relabel selected risky states with corrective actions. Across four bimanual manipulation tasks, PGDG consistently outperforms spatial-only augmentation in both simulation and zero-shot real-world transfer. On \textit{RotateBox-Pitch}, success improves from $38\%$ to $93\%$ in simulation and from $35\%$ to $82\%$ in the real world. PGDG also enables effective foundation models fine-tuning such as GR00T~\cite{nvidia2025gr00tn1openfoundation}, increasing success from $46\%$ to $77\%$. Additional results are available in our website: \href{https://cunxid.github.io/PGDG/}{https://cunxid.github.io/PGDG/}.

\end{abstract}


\begin{figure*}[!t]
  \centering
  \safeincludegraphics[width=\linewidth]{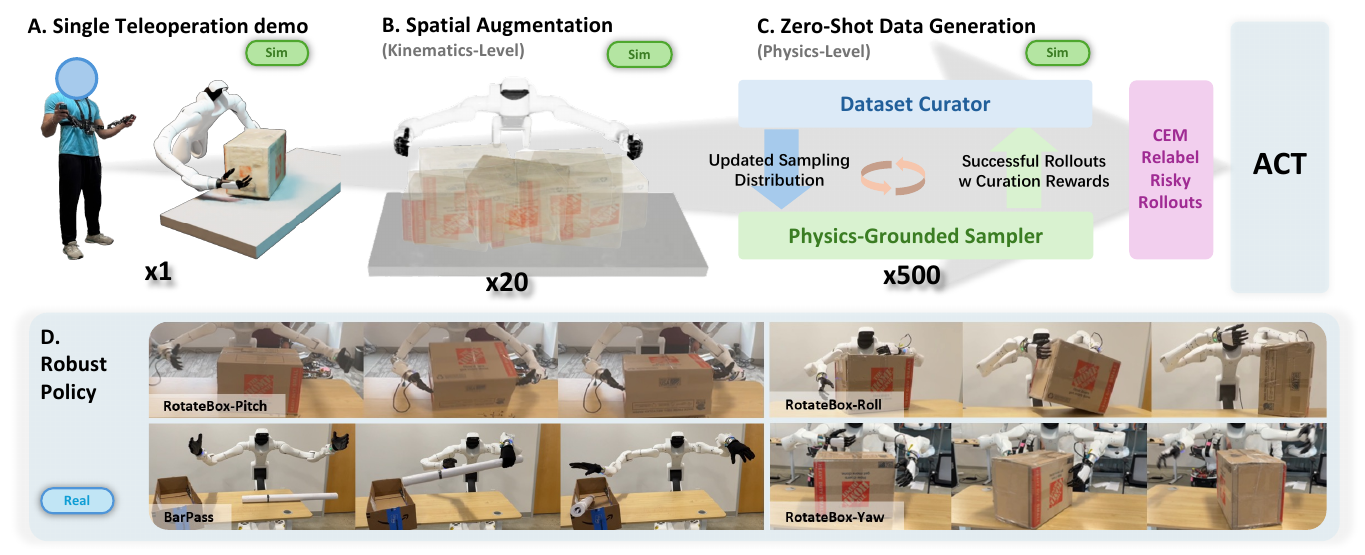}
  \caption{\textbf{One-demo $\rightarrow$ compact recovery dataset $\rightarrow$ robust policy.} From a single or few bimanual demonstrations, PGDG iteratively synthesizes a compact dataset of physically valid, diverse, and informative recovery trajectories, leading to higher success than spatial-only augmentation in simulation and on hardware.}
  \label{fig:introduction}
\end{figure*}

\section{Introduction}
\IEEEPARstart{B}{ehavior} cloning (BC) is attractive for robot manipulation due to its simplicity, scalability, and compatibility with large demonstration datasets. Yet robust performance requires not only learning nominal execution, but also learning how to recover from inevitable test-time deviations. Acquiring such recovery supervision is expensive, because it demands substantial human effort to collect demonstrations, often requiring repeated teleoperation, environment resets, and many trial-and-error rollouts to capture physically valid recovery behaviors. As a result, a naively collected dataset with limited diversity and recovery behaviors can lead BC to learn a narrow successful behavior with little corrective supervision, especially in contact-rich bimanual manipulation.

This challenge is particularly severe in contact-rich bimanual tasks, where success depends on precise coordination, contact timing, and force application. Small deviations can alter contact sequences, destabilize the object, or cause irreversible drift, making recovery highly local and physics-sensitive. Robust imitation in these settings therefore requires supervision not only for nominal execution, but also for physically valid corrective behavior near the demonstrated trajectory.

A natural response is to expand the data through augmentation, which contains two parts: \emph{generation} and \emph{curation}. On the generation side, spatial and contextual augmentation methods such as MimicGen~\cite{pmlr-v229-mandlekar23a} and DexMimicGen~\cite{jiang2025dexmimicgen}  broaden coverage across poses and scenes, but geometric similarity does not guarantee physical feasibility or recoverability in contact-rich settings. Physics-grounded methods such as PhysicsGen~\cite{yang2026physicsdrivendatagenerationcontactrich} and SPIDER~\cite{pan2026spiderscalablephysicsinformeddexterous} improve physical validity by incorporating physical constraints during generation, yielding trajectories that are more feasible for contact-rich manipulation than purely geometric augmentation. However, generation alone is not enough, because the goal is not merely to produce more trajectories, but to also curate a dataset that emphasizes informative, diverse, and task-relevant recovery supervision. On the curation side, recent methods for BC dataset curation focus on selecting or reweighting existing data~\cite{chen2025demoscore, pmlr-v305-agia25a, dass2025datamil, calian2025datarater, zhang2025scizor}, instead of generating the dataset. Moreover, these approaches typically query the policy during curation, relying on repeated policy optimization, evaluation rollouts, or learned policy-dependent scoring signals to assess data utility~\cite{chen2025demoscore,pmlr-v305-agia25a}. As a result, they often incur substantial computational overhead during curation. We therefore ask: 

\textit{For dexterous bimanual manipulation, how can one or few demonstrations be expanded, in a zero-shot manner, into a compact, physically grounded, and well-curated recovery dataset that benefits both task-specific imitation policies and broader robotics foundation models?}

To this end, we propose \textbf{PGDG (Physically Grounded Data Generation)}, a data-generation framework with zero-shot curation that expands one or a few demonstrations into a compact dataset of physically plausible, successful, and diverse recovery behaviors. Rather than treating augmentation as merely generating more trajectories, we formulate it as \textbf{generation guided by curation}. The key objective is to build a compact recovery-focused dataset, not just a larger one, by concentrating supervision on behaviors that are physically valid, still recoverable, and complementary to one another. To address this, PGDG alternates between a physics-grounded sampler that produces successful rollout candidates in simulation and a curator that prioritizes informative recoveries, filters redundant behaviors, and refines the sampling distribution toward underrepresented yet useful regions of the control space. The resulting dataset is then further improved by selectively relabeling a small number of risky states with corrective actions, providing richer supervision for robust policy learning from one or a few demonstrations. Such datasets can be used not only for single-task policy learning, but also for fine-tuning broader robotic foundation models. Our main contributions are:
\begin{itemize}
    \item \textbf{Task-agnostic physics-grounded data generation.} We propose a simulation-based pipeline that expands one or a few demonstrations into successful physics-grounded trajectories for contact-rich bimanual manipulation using only a binary success predicate for each task.
    \item \textbf{Zero-shot, policy-free dataset curation.} We curate a compact and diverse recovery dataset by prioritizing informative trajectories and removing redundancy, without policy queries during curation.
    \item \textbf{Robust policies from generated data.} Across multiple tasks, BC trained on PGDG-generated data consistently outperforms spatial augmentation alone in simulation and zero-shot real-world transfer, and the same data also improves foundation models such as GR00T~\cite{nvidia2025gr00tn1openfoundation} via fine-tuning.
\end{itemize}

\section{Related Work}

\subsection{Data Generation in Simulation}
Complementary work seeks to scale behavior cloning by generating new demonstrations from limited expert data. MimicGen~\cite{pmlr-v229-mandlekar23a} systematically adapts demonstrations across scene and object variations, and DexMimicGen~\cite{jiang2025dexmimicgen} extends this strategy to bimanual dexterous manipulation. Similar works that focus on scene randomization adopts similar technique for trajectory augmentation~\cite{pmlr-v305-ravan25a, gong2026anytaskautomatedtaskdata, pmlr-v270-garrett25a, pmlr-v305-wan25a,xue2025demogen, pmlr-v305-yu25a, pmlr-v270-hua25a, pmlr-v270-lyu25a, pmlr-v270-peng25a, pmlr-v270-kerr25a}.

However, these methods largely preserve the demonstrated motion pattern: trajectories are re-anchored to new geometries, while the underlying motion profile and contact schedule remain mostly unchanged. This is a key limitation in contact-rich bimanual manipulation, where even small pose perturbations can alter contact timing and force application, causing geometrically plausible retargeted trajectories to fail. Some physics-grounded methods such as PhysicsGen~\cite{yang2026physicsdrivendatagenerationcontactrich}, SPIDER~\cite{pan2026spiderscalablephysicsinformeddexterous} and MoMaGen~\cite{li2026momagengeneratingdemonstrationssoft} improve physical validity via physically-grounded optimization, but they primarily focus on generating feasible trajectories rather than constructing a compact, informative dataset. 

Related pipelines improve data collection efficiency by combining automation with selective human intervention, allowing humans to provide demonstrations or corrective input only in challenging regions of the task, as in intervention-based teleoperation, HITL-TAMP, and recovery-and-correction style training pipelines \cite{pmlr-v229-mandlekar23b, hu2025rac, pmlr-v267-cai25e, mandlekar2020human}. In contrast to these systems, our method does not rely on online human intervention or policy-dependent corrective feedback during data collection. Instead, it provides a lightweight and task-agnostic way to generate diverse recovery behaviors and curate the resulting trajectories with a training-free diversity objective.

\subsection{Data Curation for Behavior Cloning}
Recent work improves behavior cloning (BC) by curating demonstrations: filtering, reweighting, or selecting subsets to reduce noise and redundancy.
Demo-SCORE \cite{chen2025demoscore} leverages online rollouts to learn a demonstration filter, while CUPID \cite{pmlr-v305-agia25a} uses influence estimation to prioritize trajectories that improve downstream policy return. DataMIL \cite{dass2025datamil} uses datamodels to guide policy-driven selection, and DataRater \cite{calian2025datarater} meta-learns data values for training efficiency. SCIZOR \cite{zhang2025scizor} targets transition-level pruning and deduplication via learned progress/representation signals. Overall, these methods primarily select among existing data and typically depend on a policy and/or learned scoring models during curation.
In contrast, we focus on zero-shot synthesis: generating additional successful recovery rollouts from one/few demonstrations via trajectory search, then applying training-free diversity curation prior to policy learning.

\subsection{Recovery Supervision in Imitation Learning}

A separate line of work studies what kind of data enables robust BC under compounding error. Prior studies show that offline imitation learning is highly sensitive to dataset quality and coverage \cite{pmlr-v164-mandlekar22a}. Zhang et al.\ \cite{zhang2025actionchunking} further show that exploratory augmentation can improve closed-loop stability in continuous control. Closely related work emphasizes the importance of recovery-focused supervision: RaC \cite{hu2025rac} argues that standard demonstrations underrepresent failures and corrections, while classic interactive imitation methods such as DAgger \cite{ross2011dagger} mitigate covariate shift by labeling states from the learner's induced distribution. Other offline IL methods improve robustness through auxiliary data or learned model rollouts rather than explicit recovery synthesis \cite{ghosh2025roida,zhang2023dmil}. Our approach targets the same goal---recoverable behavior---but in a zero-shot, pre-training setting: we explicitly synthesize physics-validated recovery trajectories around the expert manifold and retain diverse successful modes through diversity-preserving selection.

% ------------------------ Preamble requirements ------------------------
% ============================================================
% Method
% ============================================================

\begin{figure*}[t]
  \centering
  \safeincludegraphics[width=\linewidth]{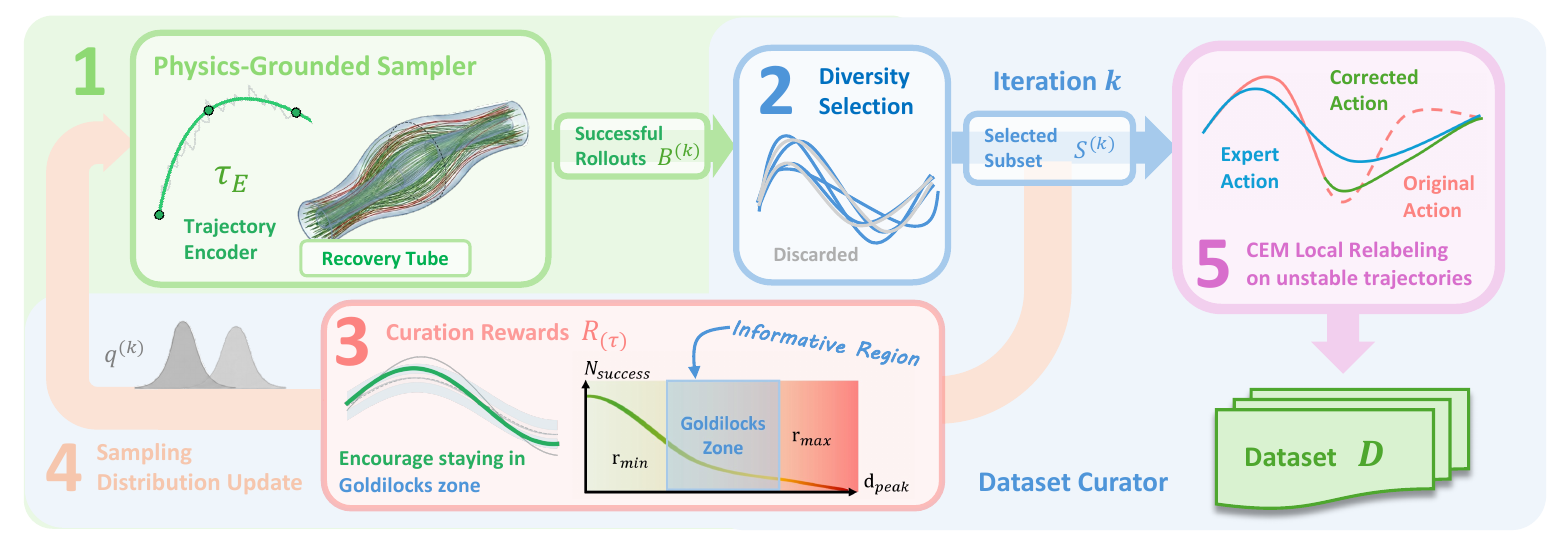}
  \caption{\textbf{Method Overview.} Starting from a single demonstration variant $\tau_E$, PGDG iterates between rollout generation and dataset curation:
\textbf{(1) Success Sampling:} sample physically feasible nominal plans with a control-point parameterization and execute rollouts in simulation.
\textbf{(2) Curation Reward:} score successful rollouts to prioritize informative recovery regions.
\textbf{(3) Diversity Selection:} remove redundant trajectories via DPP in a DCT embedding space.
\textbf{(4) Sampling Distribution Update:} refit the sampling distribution using the selected subset to encourage future exploration in informative regions.
\textbf{(5) Local Relabeling:} relabel a small number of risky states with short-horizon CEM under a full-episode success check.}
  \label{fig:method_overview}
\end{figure*}

\section{Method}
\label{sec:method}

% -------- color handles for the two components --------
% Change these two lines to any colors you like.
\definecolor{SamplerColor}{RGB}{118,196,91}
\definecolor{CuratorColor}{RGB}{46,134,193}

% Optional text macros for consistent styling in the paper/figures.
\newcommand{\sampler}[1]{\textcolor{SamplerColor}{#1}}
\newcommand{\curator}[1]{\textcolor{CuratorColor}{#1}}

An overview of our method is shown in Fig.~\ref{fig:method_overview}: we first apply spatial randomization to a demonstration to obtain multiple configuration variants. For each variant, the \sampler{\textbf{physically grounded sampler}} generates full-episode rollout candidates in simulation, while the \curator{\textbf{dataset curator}} scores successful rollouts, removes redundant trajectories, and updates the sampling distribution toward informative recovery regions. After the final curation step, we selectively relabel a small number of risky states with short-horizon CEM and add the resulting trajectories to the dataset.
\subsection{Problem Setup}
We consider deterministic dynamics
\begin{equation}
s_{t+1} = f(s_t, a_t),
\end{equation}
where $s_t\in\mathbb{R}^{d_s}$, $a_t\in\mathbb{R}^{d_a}$, and episodes have horizon $T$.
An expert demonstration is
\begin{equation}
\tau_E = \{(s^E_t, a^E_t)\}_{t=0}^{T-1} \ \cup\ \{s^E_T\}.
\end{equation}
We assume access to a simulator that executes full-episode rollouts and evaluates a binary task success predicate $\mathrm{Succ}(\tau)\in\{0,1\}$.

\subsection{Spatial Randomization}
Starting from a single expert demonstration $\tau_E$, we first apply MimicGen-style spatial randomization~\cite{pmlr-v229-mandlekar23a} to generate a set of configuration variants as shown in Fig.~\ref{fig:method_overview}~(1), which we later use as a baseline for comparison:
\[
\{\tau_E^{(i)}\}_{i=1}^{N_{\mathrm{var}}},
\]
where each $\tau_E^{(i)}$ preserves the demonstrated object-centric manipulation pattern while varying the initial robot--object geometry. Concretely, let $T^{E}_{\mathrm{ee},t}$ denote the demonstrated end-effector pose at time $t$, and let $T^{E}_{\mathrm{obj},0}$ and $T^{R,(i)}_{\mathrm{obj},0}$ denote the initial object pose in the original demonstration and in the $i$-th randomized scene, respectively. We re-anchor the demonstrated motion to the perturbed scene by expressing the trajectory relative to the object frame:
\begin{equation}
T^{R,(i)}_{\mathrm{ee},t}
\;=\;
T^{R,(i)}_{\mathrm{obj},0}
\left(T^{E}_{\mathrm{obj},0}\right)^{-1}
T^{E}_{\mathrm{ee},t},
\qquad t=0,\dots,T.
\label{eq:spatial_rand_summary}
\end{equation}
Executing the re-anchored trajectory in the perturbed scene yields a configuration variant $\tau_E^{(i)}$. Repeating this process produces multiple variants that broaden coverage over relative robot--object poses while preserving the demonstrated manipulation skill. We also added a short blending trajectory to avoid an abrupt jump from the reset state to the re-anchored motion, the details are elaborated in Sec.~\ref{app:spatial_randomization}. For each variant $\tau_E^{(i)}$, we then apply our method, shown in Fig.~\ref{fig:method_overview}.
\subsection{Overview of Generation and Curation}
\label{sec:curation_framework}

In contact-rich bimanual manipulation, useful supervision should not only cover nominal execution, but also \emph{recoverable deviations} around the expert manifold. Our goal is twofold: first, to focus data generation on the informative region, where deviations are non-trivial yet still recoverable; and second, to ensure that the final dataset covers distinct recovery modes rather than many near-duplicate trajectories.

Concretely, at iteration $k$ as shown in Fig.~\ref{fig:method_overview}, the sampler uses the current proposal distribution $q^{(k)}$ to generate a candidate set $\mathcal{B}^{(k)}$ of successful rollout trajectories in physical simulation. The curator then retains only successful rollouts and evaluates them using two complementary criteria: a curation reward $R^{(k)}(\tau)$ that measures whether a trajectory visits informative and still recoverable regions, and a diversity objective $\mathrm{Div}(S^{(k)})$ that favors non-redundant coverage of recovery behaviors. At a high level, this tradeoff can be summarized as
\[
\begin{aligned}
\max_{S^{(k)} \subseteq \mathcal{B}^{(k)}} \quad
& \sum_{\tau \in S^{(k)}} R^{(k)}(\tau)
\;+\;
\lambda\,\mathrm{Div}\!\left(S^{(k)}\right) \\
\text{s.t.}\quad
& \mathrm{Succ}(\tau)=1,\qquad \forall \tau \in S^{(k)}, \\
& \tau \text{ is obtained by rollout in physical simulation.}
\end{aligned}
\]
This objective should be understood as a conceptual description of the curator and is implemented \emph{hierarchically} rather than optimized in a single joint step: we first evaluate successful rollouts using $R^{(k)}(\tau)$ to characterize their informativeness and recoverability, and then enforce diversity through subset selection over the retained candidates. The detailed formulation of the curation reward is given in Eq.~\eqref{eq:tube_reward}, while the diversity term is instantiated in Eq.~\eqref{eq:div_dpp}.

The selected subset $S^{(k)}$ is then used to refine the proposal distribution for the next iteration:
\begin{equation}
q^{(k+1)} \leftarrow \mathrm{Update}\!\left(q^{(k)}, S^{(k)}\right).
\label{eq:sampler_update_overview}
\end{equation}
The role of this update is not merely bookkeeping: it feeds the curator's selection back into the sampler, shifting future rollouts toward regions of the control space that repeatedly yield successful, informative, and non-redundant recovery behaviors. In this way, generation and curation form a closed loop that progressively concentrates exploration on what we call the ``Goldilocks Zone'' in Fig.~\ref{fig:method_overview}. We next describe how the \sampler{sampler} generates rollout candidates, after which the \curator{curator} scores, selects, updates, and relabels them.

\subsection{\sampler{Physically Grounded Sampler}}
\label{sec:pgf}

The physically grounded \sampler{sampler} generates nominal plans, executes full-episode rollouts in simulation, and keeps only successful trajectories. Its role is to produce physically valid candidate recoveries that can subsequently be scored and curated by the dataset \curator{curator}.

\subsubsection{Trajectory Parameterization and Sampling Distribution}

Directly searching over the full action sequence $a_{0:T-1}$ is high-dimensional. We therefore parameterize a nominal plan using $M$ control points $C \in \mathbb{R}^{M \times d_a}$ together with a fixed decoder $G$:
\begin{equation}
a^{\mathrm{nom}}_{0:T-1} = G(C), \qquad c=\mathrm{vec}(C)\in\mathbb{R}^{M d_a}.
\end{equation}
At iteration $k$, the sampler draws control-point parameters from a diagonal Gaussian proposal distribution,
\begin{equation}
q^{(k)}(c)=\mathcal{N}\!\left(\mu^{(k)},\mathrm{diag}\!\big((\sigma^{(k)})^2\big)\right),
\label{eq:qdist}
\end{equation}
where $\mu^{(k)}$ and $\sigma^{(k)}$ denote the mean and standard deviation of the control-point vector at iteration $k$. This particular parameterization is not essential to our framework. The subsequent curation, diversity selection, and relabeling stages only require the ability to sample candidate control-point parameters, so $q^{(k)}$ can be replaced by other proposal families without changing the rest of the pipeline. For example, when stronger multimodal coverage is desired, one may substitute a richer family such as a Gaussian mixture model~\cite{bishop2006prml}.

At iteration $k$, we sample control-point vectors $c_n^{(k)} \sim q^{(k)}(c)$, decode each sample into a nominal plan $a^{\mathrm{nom},(n)}_{0:T-1}=G(c_n^{(k)})$, and execute the plan in simulation to obtain rollout trajectories $\tau_n^{(k)}$. We then keep only successful rollouts and define
\[
B^{(k)} \triangleq \{\tau_n^{(k)} : \mathrm{Succ}(\tau_n^{(k)})=1\}.
\]
The resulting set $B^{(k)}$ is passed to the dataset curator for reward evaluation, diversity selection, and sampler update.

\subsection{\curator{Dataset Curator}: Curation Reward}
\label{sec:curation_reward}

We summarize trajectory difference by the maximum distance to the expert manifold, because the largest deviation is the most relevant indicator of how far a rollout departs from nominal behavior before recovery. We measure distance to the expert \emph{state manifold} via nearest neighbors in a task-relevant subspace $\psi(\cdot)$ (specified in Table.~\ref{tab:pgdg_policy_params})~\cite{ross2011dagger} to tolerate timing drift (lag/catch-up):
\begin{equation}
d(s)\triangleq \min_{j\in\{0,\dots,T\}} \left\| \psi(s)-\psi(s^E_j)\right\|_2.
\label{eq:manifold_dist}
\end{equation}
For a rollout $\tau=\{s_t\}_{t=0}^{T}$, let $d_t(\tau)\triangleq d(s_t)$ and define the peak deviation:
\begin{equation}
d_{\mathrm{peak}}(\tau) \triangleq \max_{t\in\{0,\dots,T\}} d_t(\tau).
\label{eq:dpeak}
\end{equation}

We define a recovery tube from the distribution of successful rollouts in the current iteration. Specifically, the outer tube radius is set as a high quantile of successful peak deviations,
\begin{equation}
r_{\max}^{(k)} \triangleq \mathrm{Quantile}_{q_{\max}}\Big(\{d_{\mathrm{peak}}(\tau_n): \mathrm{Succ}(\tau_n)=1\}\Big),
\label{eq:rmax_select}
\end{equation}
and the inner tube radius is set as a low quantile of the same distribution,
\begin{equation}
r_{\min}^{(k)} \triangleq \mathrm{Quantile}_{q_{\min}}\Big(\{d_{\mathrm{peak}}(\tau_n): \mathrm{Succ}(\tau_n)=1\}\Big),
\label{eq:rmin_select}
\end{equation}
where \(\mathrm{Quantile}_{q}\) denotes the value below which a fraction \(q\) of successful rollouts fall. Intuitively, \(r_{\min}^{(k)}\) excludes near-demo trivial successes, while \(r_{\max}^{(k)}\) captures the typical outer edge of recoverable behavior without being dominated by a single extreme rollout. The tube is adaptive because both radii are recomputed at each iteration from the current successful batch, allowing the target recovery band to expand or contract as the sampler improves.

Using \((r_{\min}^{(k)}, r_{\max}^{(k)})\), we define a curation reward that favors trajectories whose deviations stay within the informative recovery band:
\begin{equation}
\begin{split}
R^{(k)}(\tau) \triangleq \frac{1}{T+1}\sum_{t=0}^{T}
\Bigl[
1
-\mathrm{ReLU}(r_{\min}^{(k)}-d_t(\tau)) \\
-\mathrm{ReLU}(d_t(\tau)-r_{\max}^{(k)})
\Bigr].
\end{split}
\label{eq:tube_reward}
\end{equation}
This reward is highest when the trajectory remains inside the "Goldilocks zone" shown in Fig.~\ref{fig:method_overview}, and decreases when it stays too close to the demonstration or strays beyond the recoverable range. As a result, trajectories receive higher scores when they spend more time in the informative region between trivial replay and unrecoverable failure.

\subsection{\curator{Dataset Curator}: Diversity Selection and Sampler Update}
\label{sec:diversity_update}

The dataset \curator{curator} next enforces dataset-level diversity among successful rollouts. It embeds trajectories with discrete cosine transform (DCT) features, selects a compact subset via determinantal point process (DPP), and updates $q^{(k)}$.

\subsubsection{Trajectory Embedding}
After the \sampler{sampler} returns successful and informative recovery trajectories, the dataset \curator{curator} focuses on \emph{dataset-level diversity} among the retained candidates. To compare successful rollouts directly, we represent each trajectory using its task-relevant state-action evolution rather than its distance to the expert demonstration.

For a successful rollout \(\tau=\{(s_t,a_t)\}_{t=0}^{T-1}\cup\{s_T\}\), we form per-step features $x_t$, and stack over time to obtain the rollout feature sequence $X(\tau)$:
\begin{equation}
x_t \triangleq [\psi(s_t); a_t], \quad X(\tau) \triangleq [x_0, x_1, \dots, x_{T-1}].
\label{eq:per_step_feature}
\end{equation}
where \(\psi(s_t)\) denotes the task-relevant state representation at time \(t\), and \(a_t\) is the executed action.
To obtain a compact trajectory descriptor, we apply DCT along the temporal dimension of \(X(\tau)\) and retain the low-frequency coefficients to form an embedding \(\phi(\tau)\) (Sec.~\ref{app:dct}). This embedding summarizes the overall state-action evolution of the rollout while remaining compact and training-free.

One design choice we make is to use peak deviation to label recoverability, but DCT to label diversity. Recoverability is governed primarily by the most critical off-manifold state visited during a rollout. Diversity, in contrast, depends on the overall temporal pattern of how a trajectory departs from and returns to the expert manifold, so we use a whole-trajectory representation to preserve distinct recovery modes and remove redundancy.
\subsubsection{DPP Selection}
\begin{figure}[t]
  \centering
  \safeincludegraphics[width=\linewidth]{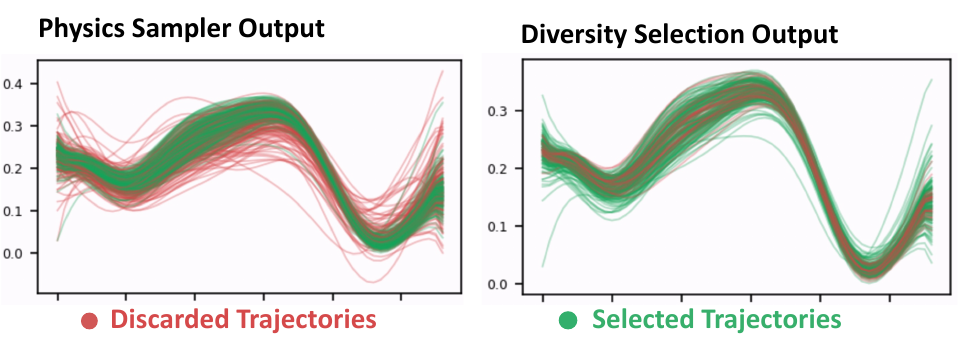}
  \caption{Data generation example. Left: all sampled rollouts/trajectories generated by the \sampler{sampler}; green indicates successful trajectories and red denotes failed ones. Right: the subset of selected trajectories retained and stored in the dataset by the \curator{curator}; the red ones are discarded due to excessive similarity under DPP selection.}
  \label{fig:data_example}
\end{figure}

We instantiate the diversity term in Sec.~III-C using a DPP set function:
\begin{equation}
\mathrm{Div}(S) \triangleq \log \det (L_S + \epsilon I),
\label{eq:div_dpp}
\end{equation}
where \(L_S\) is the principal submatrix of the DPP kernel associated with subset \(S\), and \(\epsilon > 0\) is a small constant for numerical stability. We then select a subset of \(m\) successful trajectories by maximizing $\mathrm{Div}(S)$, which favors trajectories that are both successful and diverse in the embedding space, more details are elaborated in Sec.~\ref{app:dpp}. Intuitively, this step removes near-duplicate recoveries and preserves a compact set of trajectories that covers different recovery modes. One representative data generation example is shown in Fig.~\ref{fig:data_example}.

\subsubsection{Updating the Sampling Distribution}
The selected set $S^{(k)}$ identifies regions of the control-point space that repeatedly yield successful, informative, and non-redundant recovery behaviors. We therefore use $S^{(k)}$ to refit the proposal distribution for the next iteration. Let $c_i \in \mathbb{R}^{M d_a}$ denote the sampled control-point parameter associated with trajectory $\tau_i \in S^{(k)}$, and let $w_i \ge 0$ be a scalar weight monotone in the curation score $R^{(k)}(\tau_i)$. We model the proposal distribution as a Gaussian over the full control-point vector,
\[
q^{(k)}(c)=\mathcal{N}\!\bigl(\mu^{(k)}, \Sigma^{(k)}\bigr),
\]
where $\mu^{(k)} \in \mathbb{R}^{M d_a}$ is the mean control-point vector at iteration $k$, and $\Sigma^{(k)} \in \mathbb{R}^{M d_a \times M d_a}$ is the covariance matrix that determines the spread of the sampler in the control-point space. In this paper, for simplicity and robustness, we instantiate $\Sigma^{(k)}$ as a diagonal covariance. We then update the proposal by weighted moment matching over the retained control points:
\begin{equation}
\mu^{(k+1)}
=
\frac{\sum_{i\in S^{(k)}} w_i c_i}{\sum_{i\in S^{(k)}} w_i + \varepsilon},
\label{eq:update_mu_main}
\end{equation}
\begin{equation}
\Sigma^{(k+1)}
=
\frac{\sum_{i\in S^{(k)}} w_i (c_i-\mu^{(k+1)})(c_i-\mu^{(k+1)})^\top}{\sum_{i\in S^{(k)}} w_i + \varepsilon}
+\delta I,
\label{eq:update_sigma_main}
\end{equation}
where $\varepsilon>0$ avoids division by zero, $\delta>0$ is a small covariance floor that prevents premature collapse of exploration, and $I$ is the identity matrix. In our implementation, only the diagonal entries of $\Sigma^{(k+1)}$ are retained, so the sampler remains a diagonal Gaussian. 

Across iterations, this update gradually shifts probability mass from broad exploration toward regions of control space that repeatedly yield successful and informative recovery behaviors. Early iterations cover many perturbations, including trivial successes and outright failures; later iterations concentrate on higher-value regions while retaining enough variance to continue exploring nearby alternatives.

\subsection{\curator{Dataset Curator}: Selective Local Relabeling with CEM}
\label{sec:relabel}
Although the curation reward encourages generated trajectories to remain in the informative recoverable region, the ``Goldilocks Zone'', it does not impose a hard constraint that every visited state must stay safely inside that band. As a result, even successful and selected trajectories can still contain a small number of locally risky states shown in Fig.~\ref{fig:CEM_example}. In contact-rich bimanual manipulation, such states are precisely where the executed actions are most likely to be jerky or unnecessarily unstable, even when the overall rollout eventually succeeds. We therefore selectively use Cross Entropy Method (CEM) relabeling not to regenerate whole trajectories, but to refine local supervision at the few states where corrective behavior matters most.

\subsubsection{Which States to Relabel}
We relabel the top-$K_{\mathrm{rel}}$ most risky states across the entire curated dataset. Specifically, over all states in all selected trajectories $\tau \in S^{(k)}$, we use the distance-to-expert score $d_t(\tau)$ defined earlier as the risk measure, and select the top-$K_{\mathrm{rel}}$ states with largest values. We additionally enforce a minimum temporal separation within each trajectory to avoid redundant relabels around the same deviation peak.

\subsubsection{Relabeling with Full-Episode Success Check}
For each selected relabel point \((\tau,t)\in \mathcal{T}_{\mathrm{rel}}^{(k)}\), let \(\bar u_{0:H-1}\) be the reference action segment from trajectory \(\tau\) starting at time \(t\). We optimize an action sequence \(u_{0:H-1}\) over a horizon of \(H\) steps, but evaluate it using a \emph{full} continuation rollout: we first apply \(u_{0:H-1}\), then resume the reference plan for the remaining steps. This encourages locally corrective actions that still lead to eventual task success, rather than locally convenient but globally inconsistent fixes.

For a candidate sequence \(u\), we define
\begin{equation}
\begin{aligned}
J(u; s_t) =\;&
w_{\mathrm{fail}}\cdot \mathbf{1}\!\left[\neg \mathrm{Succ}\!\left(\mathrm{Rollout}(s_t, u_{0:H-1}, \bar u_{H:\,T-t-1})\right)\right] \\
&+ w_{\mathrm{tube}}\sum_{h=1}^{H}\mathrm{ReLU}\!\big(d(s_{t+h})-r_{\max}^{(k)}\big)^2 \\
&+ w_{\mathrm{ref}}\sum_{h=0}^{H-1}\|u_h-\bar u_h\|_2^2 ,
\end{aligned}
\label{eq:mppi_cost}
\end{equation}
where the first term enforces full-episode success, the second penalizes leaving the recoverable tube, and the third keeps the correction close to the original nominal action segment.

CEM returns an optimized sequence \(u^{\star}_{0:H-1}\) at \(s_t\), which we store as the corrective target
\begin{equation}
y_t = (u^{\star}_0,\dots,u^{\star}_{H-1}),
\end{equation}
paired with the observation \(o_t\) at \(s_t\). We store the full horizon \(H\) so that the relabeled target matches exactly the sequence validated by the full-episode success check.

\begin{figure}[t]
  \centering
  \safeincludegraphics[width=\linewidth]{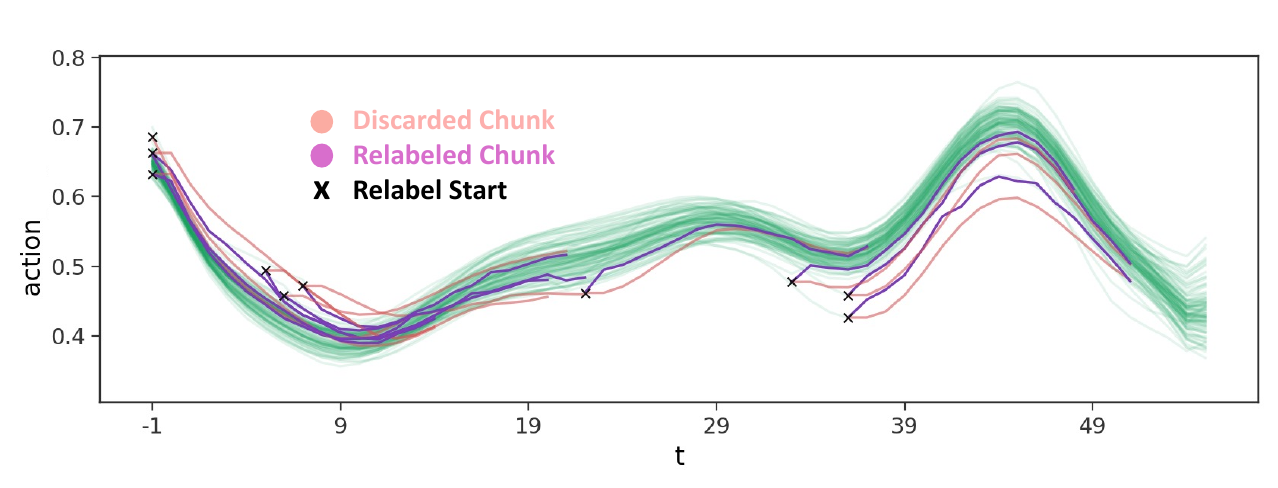}
\caption{Selective local relabeling with CEM. For a selected off-manifold state (black \texttimes), we optimize a short corrective action chunk (purple) that replaces the original executed segment (red), while preserving the surrounding trajectory context. This improves local action supervision at risky states without relabeling the entire rollout.}  \label{fig:CEM_example}
\end{figure}

We perform relabeling after the sampler update because the two stages serve different roles: sampler adaptation operates at the rollout level over global control-point parameters, whereas relabeling refines only local, state-conditional supervision. Keeping them separate preserves a clean distinction between global proposal adaptation and local action correction. A representative example is shown in Fig.~\ref{fig:CEM_example}.

In this way, the final dataset is not only diverse and successful at the trajectory level, but also provides corrective supervision at the states where failures are most likely to occur.

% -------------------Policy Training-----------------------
\section{Policy Training}\label{sec:policy_training}

In this work, we employed ACT~\cite{zhao2023learningfinegrainedbimanualmanipulation} and GR00T~\cite{nvidia2025gr00tn1openfoundation} as our BC models. We show that our data generation method is policy-agnostic. The models are trained on depth images with high-low byte encoding to ensure accuracy. For the ACT model, we used ResNet-18~\cite{he2015deepresiduallearningimage} as the vision backbone.

The output of \algo\ is a dataset of successful trajectories, optionally augmented with locally relabeled corrective actions at selected risky states. We store the resulting training data as a set of timestep-level observation--action pairs,
\[
\mathcal{D}=\{(o_t,a_t)\},
\]
extracted from these trajectories, where \(o_t\) denotes the full observation at timestep \(t\) and \(a_t\) the executed action. Each observation \(o_t\) contains both a depth image and the robot proprioceptive state. We denote by \(\mathbf I_t\) the depth image extracted from \(o_t\), and by \(\mathbf S_t\) the proprioceptive state extracted from \(o_t\). Pairs drawn from the same trajectory retain their temporal order, so for policy training we can form a state-history input \(\mathbf S_{t-H_s+1:t}\) by collecting the proprioceptive components from the most recent \(H_s\) observations in the same trajectory. These ordered samples are then regrouped into supervision targets according to the policy architecture.

For ACT, each training example is constructed by taking the current depth image \(\mathbf I_t\) together with a short history of proprioceptive states \(\mathbf S_{t-H_s+1:t}\) as input, where \(H_s\) denotes the length of the state-history window. The supervision target is a future action chunk
\begin{equation}
a^{\mathrm{tar}}_{t:t+k-1}
\triangleq
(a_t, a_{t+1}, \dots, a_{t+k-1}),
\label{eq:policy_target_chunk}
\end{equation}
where \(k\) is the action chunk length. The policy predicts
\begin{equation}
\tilde{\mathbf a}_{t:t+k-1}
=
\pi_\theta\!\left(
\mathbf I_t,
\mathbf S_{t-H_s+1:t},
\mathbf z
\right),
\label{eq:act_policy_main}
\end{equation}
where \(\mathbf z\) is the latent variable used in the CVAE objective.

For GR00T fine-tuning, we use the same trajectory dataset, but convert it into the supervision format required by GR00T. As with ACT, the model takes inputs derived from \(o_t\) and predicts future actions extracted from the same successful trajectory:
\begin{equation}
\tilde{\mathbf a}_{t:t+k-1}
=
\pi_\theta\!\left(
\mathbf I_t,
\mathbf S_t
\right).
\label{eq:groot_policy_main}
\end{equation}
GR00T uses a different action chunk length and input formatting, details shown in Table.~\ref{tab:params}.

Across both policies, depth is encoded by high--low bytes to preserve metric precision. We primarily use the current depth image as visual input, and optionally append the first image of the episode as an additional progress cue. We also apply depth preprocessing and mild augmentation for sim-to-real transfer, including clipping, cropping, and simulated sensor artifacts; these details are deferred to Appendix~\ref{ACT_appendix_depth}. State history is included because depth-only observations can be temporally ambiguous in contact-rich manipulation, the exact construction is given in Appendix~\ref{ACT_appendix_input}.

% a. too wordy? b. need adjust the order of these contents and formulas c. need proofread tmr

% Real world deployment
% 1. depth image is aligned with left, sides are cropped, single channel, stereo matching
% 2. contact force sensor to prevent overshooting since visual clue is not always strong, 30% less N*m, a potential table here

% haoran:
% ACT training details

% -------------------Experiments-----------------------

\section{Behavior Cloning Experiments}
\subsection{Hardware Setup}
\begin{figure}[t]
  \centering
  \safeincludegraphics[width=\linewidth]{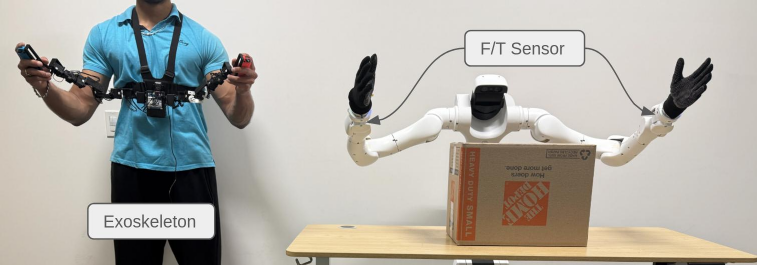}
  \caption{Teleoperation setup. We utilized a GELLO-style exoskeleton to perform cartesian space teaching in simulation.}
  \label{fig:teleoperation}
\end{figure}

We collect the source demonstration on the  bimanual mobile manipulator with a pair of 7-DoF arms and wrist F/T sensors. To obtain a high-quality source trajectory demo with minimal operator burden, we utilized a wearable GELLO-style~\cite{10801581} exoskeleton to perform end-effector (EE) space teaching, a picture of the setup is shown in Fig.~\ref{fig:teleoperation}.

\subsection{Bimanual Tasks and Parameters}

We selected $4$ bimanual tasks to benchmark policy performance: \textsc{RotateBox-Roll}, \textsc{RotateBox-Pitch}, \textsc{RotateBox-Yaw}, and \textsc{BarPass}. Across all RotateBox tasks, the manipulated object is a medium-sized cardboard box with outer dimensions $24\text{ in}\times 18\text{ in}\times 12\text{ in}$. We define success by axis-aligned orientation accuracy: letting $\theta_t^{(k)}$ denote the current box
rotation angle about axis $k\in\{\text{roll},\text{pitch},\text{yaw}\}$ and $\theta_{\mathrm{des}}^{(k)}$ the desired angle, an episode
is successful if the binary success predicate is $1$, calculated as
\begin{equation}
\mathrm{Succ}(\tau)_{\text{RotateBox}} = \left\lVert \theta_T^{(k)} - \theta_{\mathrm{des}}^{(k)} \right\rVert_2 < \epsilon_\theta,
\end{equation}
where $T$ is the final time step and $\epsilon_\theta$ is a small completion tolerance. Similarly for Barpass, we define success by terminal pose accuracy in position and yaw only. 
Let $p_t^{\text{bar}}\in\mathbb{R}^3$ be the bar position and $\psi_t$ be the bar yaw angle (rotation about the world $z$ axis), with desired targets $(p_{\text{des}}^{\text{bar}}, \psi_{\text{des}})$. 
An episode is successful if
\begin{equation}
\begin{aligned}
\mathrm{Succ}(\tau)_{\text{BarPass}} =
&\Big(\lVert p_T^{\text{bar}} - p_{\text{des}}^{\text{bar}} \rVert_2 < \epsilon_p\Big) \\
&\wedge\ \Big( \left\lVert \theta_T^{\text{yaw}} - \theta_{\mathrm{des}}^{\text{yaw}} \right\rVert_2 < \epsilon_\theta\Big).
\end{aligned}
\end{equation}

where $\epsilon_p$ is also completion tolerance. The randomization details and completion tolerance are specified in Table~\ref{tab:domain_rand}. 

\begin{table}[t]
\label{tab:task}
\centering
\caption{Domain randomization for Bi-manual Tasks}
\label{tab:domain_rand}
\setlength{\tabcolsep}{3.5pt}
\renewcommand{\arraystretch}{1.05}
\footnotesize
\begin{tabularx}{\columnwidth}{l>{\centering\arraybackslash}X>{\centering\arraybackslash}X}
\toprule
\textbf{Parameter} & \textbf{RotateBox} & \textbf{BarPass} \\
\midrule
Init.\ obj.\ trans.\ pert.\ (cm) & $[\pm8,\ \pm8,\ 0]$ & $[\pm5,\ \pm5,\ 0]$ \\
Init.\ obj.\ rot.\ pert.\ (rad)  & $[0,\ 0,\ \pm0.3]$      & $[0,\ 0,\ \pm0.3]$ \\
Object mass (kg)                 & $[0.5,\ 3.0]$           & $[0.25,\ 0.75]$ \\
Friction coefficients             & $[0.8,\ 1.2]$           & $[0.2,\ 0.4]$ \\
Task horizon (s)                 & $3.6$(r) $2.8$(p) $3.05$(y)    &  $11.4$\\
Completion. tol.                & $\epsilon_{\theta}=0.1$(r, p) $0.2$(y)  & $\epsilon_p=0.3, \epsilon_{\theta}=0.2$\\

\bottomrule
\end{tabularx}
\end{table}

\subsection{Data Generation Overview}
We run $5$ iterations of dual-loop generation for each spatially randomized source demo. The DPP selection step omitted 11.46\% of the feasible trajectory for \textsc{RotateBox-Roll}, 14.07\% for \textsc{RotateBox-Pitch}, 21.88\% for \textsc{RotateBox-Yaw} and 16.15\% for \textsc{Bar}.

\subsection{Policy Evaluation in Simulation}
\begin{figure}[t]
  \centering
  \safeincludegraphics[width=\linewidth]{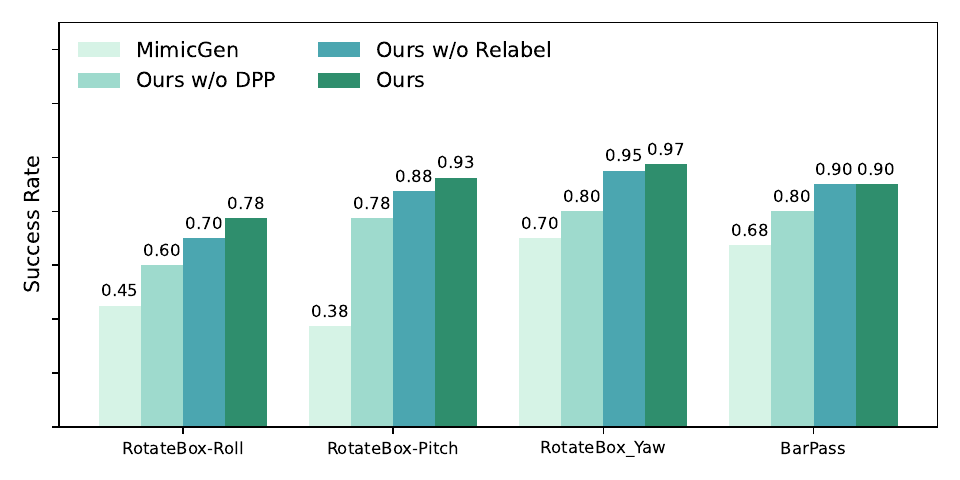}
  \caption{Policy evaluation in Simulation. \textbf{MimicGen} denotes data generated with only spatial randomization in MimicGen style. \textbf{Ours w/o Relabel} denotes data generated without CEM relabeling (Sec.~\ref{sec:relabel}). \textbf{Ours w/o DPP} denotes data generated with only inner loop sampling.}
  \label{fig:simulation_eval}
\end{figure}

Starting from a single human demonstration, \algo\ generates a dataset of thousands of trajectories. We evaluate policies in Isaac Sim~\cite{NVIDIA_Isaac_Sim} and train a depth-based ACT policy as described in Sec.~\ref{sec:policy_training}. At test time, we roll out each policy for 40 episodes under the same domain-randomization settings as used during data generation, and report the resulting success rate. Baseline definitions and performance are summarized in Fig.~\ref{fig:simulation_eval}.
% simulation evaluation figure
\begin{figure}[t]
  \centering
  \safeincludegraphics[width=\linewidth]{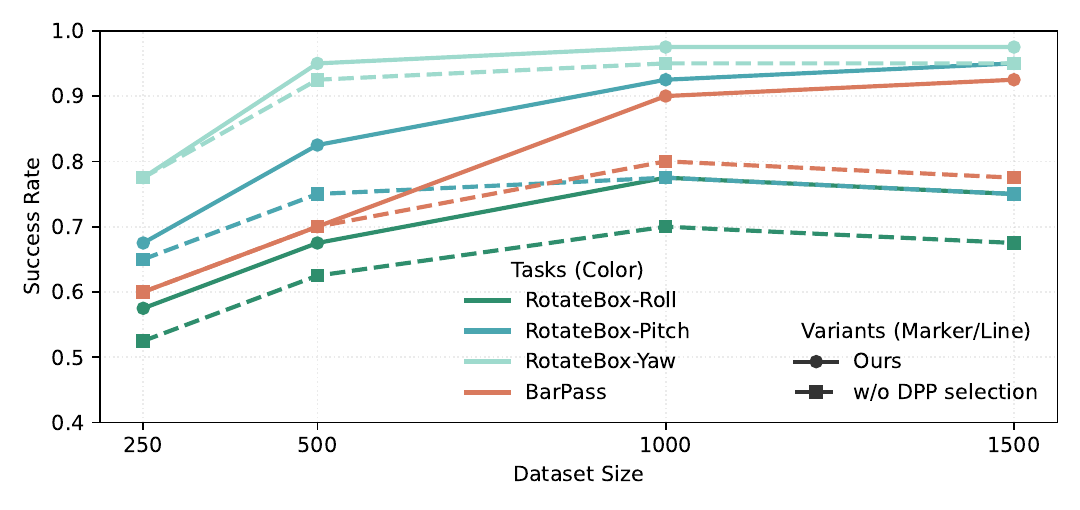}
  \caption{\textbf{Policy performance vs.\ dataset size.} Success rate of depth-based ACT policies trained on datasets of increasing size generated from a single demonstration. Without diversity curation (w/o DPP), additional trajectories can be redundant and skew the training distribution, so larger datasets do not necessarily improve performance. In contrast, DPP-based subset selection yields a more balanced, non-redundant dataset and improves the success rate per collected trajectory.}

  \label{fig:datasize}
\end{figure}
Given the dynamics and contact-critical nature of our tasks, MimicGen-style spatial randomization often fails to transfer a demonstration without additional physics-aware correction. Although spatial re-anchoring preserves scene geometry, replaying the same trajectory does not preserve the contact schedule; consequently, small differences in low-level tracking can shift contact timing and lead to missed, weak, or unstable contacts. This issue is most severe in \textsc{RotateBox-Pitch}, where the flipping motion completes in less than $0.5\,\mathrm{s}$ and success depends on a narrow timing window. As shown in Fig.~\ref{fig:simulation_eval}, policies trained on MimicGen data degrade substantially under these perturbations, while our physics-grounded generation yields recovery trajectories that remain dynamically feasible under the same randomization. Quantitatively, our full method improves success from $0.45$ to $0.78$ on \textsc{RotateBox-Roll}, from $0.38$ to $0.93$ on \textsc{RotateBox-Pitch}, from $0.70$ to $0.97$ on \textsc{RotateBox-Yaw}, and from $0.68$ to $0.90$ on \textsc{BarPass}. The data ablations further show that these gains are not explained by dataset scale alone: removing DPP curation (\emph{Ours w/o DPP}) reduces performance on all tasks, and increasing the number of generated trajectories without diversity control does not reliably improve success because redundant rollouts bias the training distribution toward a narrow set of stable solutions. In contrast, DPP-based subset selection retains a more balanced and non-redundant recovery dataset, while selective relabeling further improves label quality at off-manifold states. Overall, the results suggest that, for fast contact-rich bimanual manipulation, robustness depends not only on spatial variation, but on generating compact datasets that preserve physically valid contact timing, diverse recovery modes, and corrective supervision.
\begin{figure*}[t]
  \centering
  \safeincludegraphics[width=\linewidth]{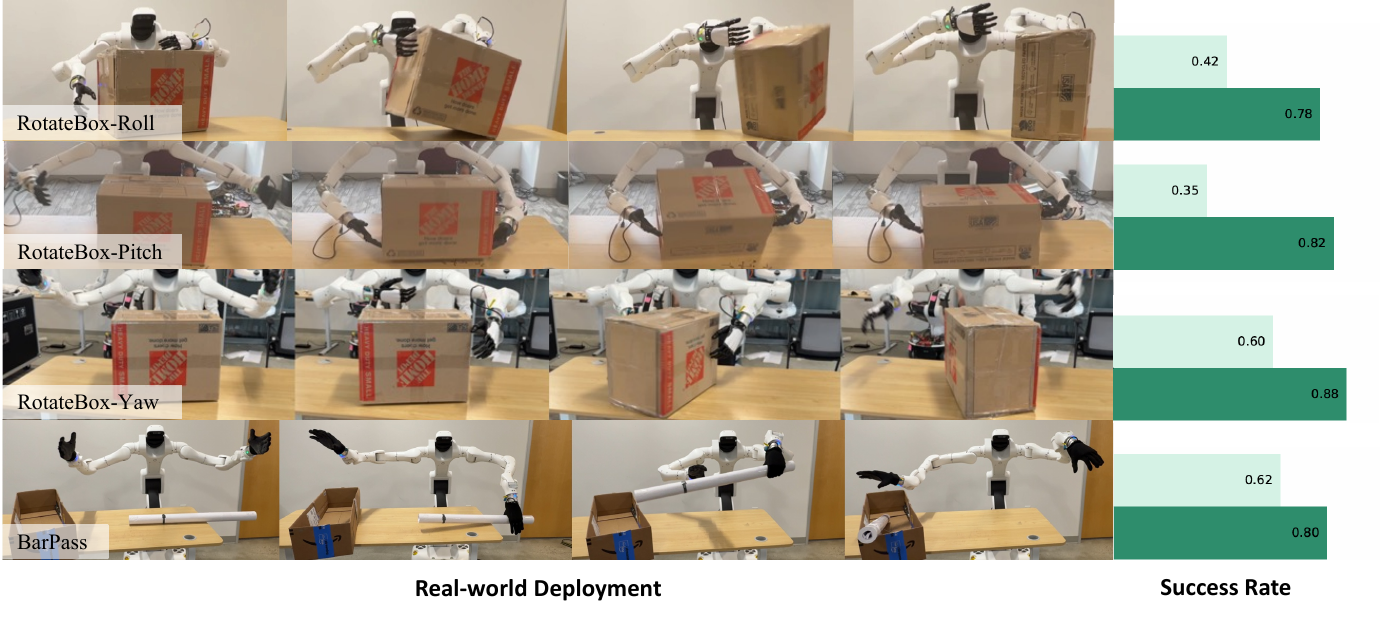}
  \caption{\textbf{Real-world experiments.} Zero-shot deployment of policies trained in simulation on four bimanual tasks. Left: representative real-world rollout frames. Right: success rate over 40 trials per task under the same initialization randomization ranges as in simulation (Tab.~\ref{tab:domain_rand}), comparing spatial-only augmentation (MimicGen) against our physics-grounded recovery data generation.}
  \label{fig:realworld}
\end{figure*}

To further validate that our generated data benefits foundation model adaptation, we fine-tune GR00T N1.6 using (i) our physics-grounded recovery dataset and (ii) MimicGen-style spatially randomized data. Table~\ref{tab:groot_ft} compares success rates after fine-tuning on the four tasks, where our data yields higher success rate consistent with our policy evaluation results.

\paragraph{Why vanilla spatial transformation fails}
Our tasks require (i) the correct \emph{contact sequence} (where and how the hands touch the object) and (ii) precise \emph{timing} in fast maneuvers. MimicGen-style spatial retargeting rigidly shifts/rotates the desired hand poses to a new object pose but largely preserves the original motion profile and contact schedule. Under pose perturbations, millimeter-level contact-point shifts or small approach-angle changes can cause missed contacts, wrong normal forces, or phase misalignment (contact occurs too early/late). In fast motions (e.g., \textsc{RotateBox-Pitch}), the low-level controller cannot correct these errors within the short contact window, so geometrically plausible retargeted trajectories still fail.

\paragraph{Role of Diversity}
We also ablate the role of diversity-based curation. Removing DPP subset selection (``Ours w/o DPP'') reduces dataset-level coverage, causing the generated trajectories to concentrate near stable solutions. Consequently, increasing the dataset size does not reliably improve performance: additional samples are often redundant and can further bias the training distribution. To quantify this effect, we evaluate policies trained on varying dataset sizes and report results in Fig.~\ref{fig:datasize}. The trend shows that, without diversity selection, larger datasets are not necessarily better; in contrast, DPP curation yields a more balanced, non-redundant dataset and improves success rate per collected trajectory.

\begin{table}[t]
\begin{center}
\caption{GR00T fine-tuning results (success rate, \%). }

\small
\setlength{\tabcolsep}{6pt}
\begin{tabular}{lcccc}
\toprule
\textbf{Fine-tuning Data} & \textsc{RB-Y} & \textsc{RB-P} & \textsc{RB-R} & \textsc{Bar} \\
\midrule
\textbf{Ours (PGDG)}      & 87.5 & 77.5 & 72.5 & 70.0 \\
\textbf{Spatial (MimicGen-Style)} & 65.0 & 40.0 & 45.0 & 35.0 \\
\bottomrule
\label{tab:groot_ft}
\end{tabular}
\end{center}

\end{table}

\subsection{Real-World Experiment}
We evaluate sim-to-real transfer by deploying the depth-based ACT policy on the real system. For each task, we run 40 trials with the same initial-state randomization ranges as in simulation (Tab.~\ref{tab:domain_rand}) and report success rates in Fig.~\ref{fig:realworld}.

Across all real-world tasks, the dominant failure mode is unexpected object slipping, which we attribute primarily to sim-to-real discrepancies in friction and contact compliance. The cardboard object deforms locally during contact---for example through surface compression and edge rounding---whereas our simulator models it as rigid. As a result, the real system exhibits lower effective friction and shifted contact normals, causing otherwise-valid contact sequences to lose stability, especially during fast transitions. Despite this gap, the hardware results remain consistent with the simulation trends: under the same initialization randomization ranges, policies trained with our physics-grounded dataset achieve higher success than those trained with spatial-only augmentation on all four tasks. These results indicate that the benefit of our method is not that it eliminates sim-to-real mismatch, but that it learns recovery behaviors that better tolerate contact uncertainty and partial execution error in real deployment.

To further validate the robustness of our trained policy, we also showed that using a 12-inch cubic box, the trained policy could robustly flip about the y-axis. The videos could be seen in our supplementary material and website.

% -------------------Limitations-----------------------
\section{Limitations and Findings}
Our data generation is fundamentally local around the provided demonstration. While the pipeline can synthesize diverse \emph{recoveries} near the expert manifold, it is not designed to discover entirely new successful modes. Therefore, if the source demonstration is suboptimal or lies in a narrow basin of success, the generated dataset may inherit this bias and miss alternative strategies that require leaving the demonstration neighborhood.
Also, the source demo needs to be collected in end-effector space.

Some interesting findings: (i) For highly dynamic-sensitive behaviors, converting demonstrated joint-space motions into Cartesian end-effector targets can degrade supervision. This mapping can be information-losing and may mismatch the robot’s Cartesian controller (e.g., unmodeled impedance/contact forces and controller-specific tracking), reducing downstream policy performance.
(ii) Larger datasets are not necessarily better: without dataset-level curation, additional trajectories can be redundant and skew the training distribution, hurting performance.
(iii) Task success is non-uniformly sensitive across joints—perturbations in a few DoFs dominate recoverability while others tolerate wider variation—suggesting recovery augmentation should be structured and state-dependent rather than uniformly applied.

\section{Conclusion}
\label{conclusion}
We presented PGDG, an iterative pipeline that turns a small number of demonstrations into a compact, diverse recovery dataset for robust behavior cloning. By alternating between a physics-grounded sampler and a dataset curator, PGDG focuses data generation on informative but recoverable deviations, preserves distinct recovery modes through diversity-based selection, and refines off-manifold supervision through selective local relabeling with full-episode success constraints.

\section*{Acknowledgments}
The author would like to thank Chaoyi Pan and for his valuable discussions and insightful feedback on the paper.  Guanya Shi holds concurrent appointments as an Assistant Professor at Carnegie Mellon University and as an Amazon Scholar. This paper describes work performed at Carnegie Mellon University and is not associated with Amazon.

\bibliographystyle{IEEEtran}
\bibliography{references}

%% Use plainnat to work nicely with natbib.

\appendices
\section{Spatial Randomziation}
\subsection{Spatial Randomization via Object-Centric Re-anchoring}
\label{app:spatial_randomization}

When actions are specified in end-effector space, we augment a single demonstration by re-anchoring the demonstrated end-effector motion to randomized object initial poses, in the style of object-centric spatial augmentation. The key idea is to preserve the demonstrated \emph{relative} manipulation skill while broadening coverage over robot--object configurations.

\paragraph{Object-centric canonicalization}
Let $T^{E}_{\mathrm{ee},t}\in SE(3)$ denote the demonstrated end-effector pose at time $t\in\{0,\dots,T\}$ in the world frame, and let $T^{E}_{\mathrm{obj},0}\in SE(3)$ be the object pose at the start of the demonstration. We express the demonstrated end-effector trajectory in the object frame at $t=0$:
\begin{equation}
\bar T_{\mathrm{ee},t}
\;\triangleq\;
\left(T^{E}_{\mathrm{obj},0}\right)^{-1} T^{E}_{\mathrm{ee},t},
\qquad t=0,\dots,T.
\label{eq:obj_centric}
\end{equation}

\paragraph{Sampling a randomized object pose}
We sample a spatial perturbation $\Delta T\in SE(3)$, typically planar translation and yaw, and define a randomized initial object pose
\begin{equation}
T^{R}_{\mathrm{obj},0} \;\triangleq\; \Delta T\, T^{E}_{\mathrm{obj},0},
\qquad
\Delta T \;=\;
\begin{bmatrix}
R_z(\Delta\theta) & \Delta p \\
0 & 1
\end{bmatrix},
\label{eq:obj_pose_rand}
\end{equation}
where $\Delta p\in\mathbb{R}^3$ (often with $\Delta p_z=0$) and $\Delta\theta\in\mathbb{R}$.

\paragraph{Re-anchoring the end-effector trajectory}
We reconstruct the randomized end-effector pose sequence by composing the same object-centric motion with the randomized object pose:
\begin{equation}
T^{R}_{\mathrm{ee},t}
\;\triangleq\;
T^{R}_{\mathrm{obj},0}\,\bar T_{\mathrm{ee},t}
\;=\;
T^{R}_{\mathrm{obj},0}\left(T^{E}_{\mathrm{obj},0}\right)^{-1} T^{E}_{\mathrm{ee},t},
\label{eq:ee_reanchor}
\end{equation}
Thus, the relative motion with respect to the object is preserved, while the absolute scene geometry is randomized.

\paragraph{Blending from the reset pose}
Let $T^{0}_{\mathrm{ee}}=(R^0,p^0)$ be the end-effector pose after environment reset in the randomized scene, and let
$T^{R}_{\mathrm{ee},0}=(R^R_0,p^R_0)$ be the first pose of the re-anchored trajectory. To avoid an abrupt jump from the reset state to the re-anchored motion, we prepend a short blending segment of length $L_{\mathrm{blend}}$:
\begin{equation}
\begin{aligned}
\alpha_\ell &\triangleq \frac{\ell}{L_{\mathrm{blend}}}, \\
p^{\mathrm{blend}}_\ell &= (1-\alpha_\ell)p^0 + \alpha_\ell p^R_0,\\
R^{\mathrm{blend}}_\ell &= \mathrm{Slerp}\!\left(R^0, R^R_0;\alpha_\ell\right),
\end{aligned}
\qquad \ell=0,\dots,L_{\mathrm{blend}}.
\label{eq:blend}
\end{equation}
The final augmented end-effector pose sequence is the concatenation of this blending segment and the re-anchored trajectory.

\paragraph{Back to actions}
Since we use cartesian pose targets as action representation, we set
\begin{equation}
a^{R}_t \leftarrow T^{\mathrm{aug}}_{\mathrm{ee},t+1}.
\end{equation}

\section{ACT Architecture}\label{ACT_appendix}

\subsection{Training Details}\label{ACT_appendix_training}

For ACT, each training sample is constructed from a successful trajectory by taking the observation at timestep \(t\) as input and the future action chunk \(a^{\mathrm{tar}}_{t:t+k-1}\) as supervision target, as defined in Sec.~\ref{sec:policy_training}. ACT is trained as a conditional variational autoencoder (CVAE). Following~\cite{zhao2023learningfinegrainedbimanualmanipulation}, a learnable classification token \(\mathbf{CLS}\) is prepended in the encoder and used to aggregate the conditioning information for predicting the latent distribution over action chunks. During training, the encoder outputs
\begin{equation}
(\mu_t,\sigma_t)
=
\mathbf{CVAE}\!\left(
\mathbf{CLS},
\mathbf{a}^{\mathrm{tar}}_{t:t+k-1},
\mathbf{S}_{t-H_s+1:t}
\right).
\end{equation}
The latent variable is then sampled via
\begin{equation}
\mathbf{z}_t=\mu_t+{\sigma_t} \odot \boldsymbol{\epsilon},
\qquad
\boldsymbol{\epsilon} \sim \mathcal{N}(\mathbf{0}, \mathbf{I}),
\end{equation}
and the policy is trained with the standard ACT objective
\begin{equation}
\mathcal{L}
=
\mathcal{L}_1(\tilde{\mathbf{a}}_{t:t+k-1}, \mathbf{a}^{\mathrm{tar}}_{t:t+k-1})
+
\lambda_{\mathrm{KL}}
D_{\mathrm{KL}}\!\left(
\mathcal{N}(\mu_t, \sigma_t^2)
\;\|\;
\mathcal{N}(\mathbf{0}, \mathbf{I})
\right).
\end{equation}
At inference time, \(\mathbf{z}_t\) is set to \(\mathbf{0}\) for deterministic execution. For relabeled samples, the target chunk has the same dimension \(k\) as standard samples.  We report the main implementation hyperparameters we used in Table~\ref{tab:params}; for additional architectural and optimization details, we refer the reader to the original ACT paper~\cite{zhao2023learningfinegrainedbimanualmanipulation}.
\begin{table}[t]
\centering
\caption{Hyperparameters and implementation details for ACT and GR00T N1.6.}
\label{tab:params}
\begin{tabular}{lcc}
\hline
\textbf{Parameter} & \textbf{ACT} & \textbf{GR00T N1.6} \\
\hline
\multicolumn{3}{l}{\textit{Policy}} \\
\hline
Optimizer        & AdamW   & AdamW \\
Learning rate    & $1\times10^{-5}$ & $1\times10^{-4}$ \\
Weight decay     & ---     & $1\times10^{-5}$ \\
Batch size       & $256$   & $64$ \\
Training steps   & $50$k   & $30$k \\
Action chunk size      & $30$           & $50$ \\
Action execution steps & $24$           & $15$ \\
No.\ of action heads   & $8$            & $32$ \\
Action head dimension  & $64$           & $48$ \\
Image history          & $[0]$          & $[0]$ \\
State history          & $[-0.75,\ 0]$  & $[0]$ \\
Control frequency      & $20$ Hz        & $20$ Hz \\
\hline
\multicolumn{3}{l}{\textit{Perception}} \\
\hline
Backbone              & ResNet-18 & Eagle-2B \\
Depth encoding        & \multicolumn{2}{c}{high--low byte (16-bit, 2-channel)} \\
Depth clipping range  & \multicolumn{2}{c}{$[0.2\,\mathrm{m},\ 1.2\,\mathrm{m}]$} \\
Image crop            & \multicolumn{2}{c}{crop $5\%$ from all boundaries} \\
\hline
\end{tabular}
\end{table}
% \begin{table}[t]
% \centering
% \caption{Policy / perception parameters.}
% \label{tab:act_params}
% \begin{tabular}{ll}
% \hline
% \textbf{Parameter} & \textbf{Value} \\
% \hline
% Action chunk size & $k=30$ \\
% Control frequency & $20$ Hz \\
% Vision backbone & ResNet-18 \\
% Depth encoding & high--low byte (16-bit, 2-channel) \\
% Depth clipping range & $[0.2\,\mathrm{m},\ 1.2\,\mathrm{m}]$ \\
% Image crop & crop $5\%$ pixels from all boundaries \\
% \hline
% \end{tabular}
% \end{table}

\subsection{Depth Image Processing}\label{ACT_appendix_depth}
The original depth images are collected at a resolution of $960 \times 600$ and resized to a $640 \times 400$ while preserving the original aspect ratio before being fed into the network. 
To preserve depth precision during training, the collected depth images are encoded using a high--low byte representation, where the first channel stores the high byte and the second channel stores the low byte, effectively yielding a 16-bit depth encoding. This representation is critical for dexterous manipulation tasks such as box flipping, where fine geometric variations in depth are necessary to accurately execute small, contact-rich motions. To better simulate real camera effects and reduce the sim-to-real gap, we apply several depth preprocessing steps. We introduce edge rounding and boundary smoothing operations to mitigate sharp discontinuities produced by idealized simulation rendering. Depth values are clipped to the range $[0.2\,\mathrm{m}, 1.2\,\mathrm{m}]$ to suppress out-of-range noise. Additionally, we crop $5\%$ of pixels from all image boundaries to account for potential dead pixels and edge artifacts commonly observed in real depth cameras.

\subsection{Policy Input}\label{ACT_appendix_input}
Specifically, state vectors from the previous $H$ timesteps are concatenated and projected into a fixed-dimensional embedding, allowing them to be jointly processed with visual, action, and latent tokens by the transformer encoder. This design is particularly important for tasks such as flipping a box using depth-only observations, where a single depth image cannot disambiguate whether the robot is approaching the object in preparation for rotation or whether the object has already been rotated. The transformer input sequence includes a learnable latent token, denoted as $\mathbf{CLS}$, which is randomly initialized and optimized jointly with the rest of the network. This token is used by the CVAE encoder during training to parameterize the latent distribution over action chunks.
\section{PGDG Algo Details}
% ============================================================
% Appendix: details moved, but NOTHING lost
% ============================================================

\label{app:algo}

\subsection{DCT embedding}
\label{app:dct}

After padding or truncating the per-step feature$x_t \triangleq [\psi(s_t); a_t] \in \mathbb{R}^d$ to a fixed horizon $\tilde T$, we stack the sequence into$
X \in \mathbb{R}^{\tilde T \times d}$.
We then apply a type-II discrete cosine transform (DCT-II) along the temporal dimension~\cite{ahmed1974discrete, pertsch2025fastefficientactiontokenization}, which yields a compact frequency-domain summary of the rollout by concentrating most of the signal energy in a small number of low-frequency coefficients. Discarding the DC component and retaining the next $K_{\mathrm{dct}}$ coefficients gives
\begin{equation}
\phi(\tau) \triangleq
\mathrm{vec}\!\left((D_{\tilde T}X)_{2:K_{\mathrm{dct}}+1,:}\right)
\in \mathbb{R}^{K_{\mathrm{dct}} d},
\end{equation}

where $D_{\tilde T}$ denotes the DCT-II matrix and $\mathrm{vec}(\cdot)$ vectorizes the retained coefficients. The resulting embedding captures the coarse state-action evolution of the rollout while filtering out high-frequency variations, making it suitable for measuring trajectory similarity in the subsequent DPP selection stage.

\subsection{DPP kernel and selection.}
\label{app:dpp}

To promote diversity and reduce redundancy, we select trajectories using a determinantal point process~\cite{kulesza2012dpp}. Given embeddings $\phi(\tau_i)$, we define an RBF kernel
\begin{equation}
L_{ij}=\exp\!\left(-\frac{\|\phi(\tau_i)-\phi(\tau_j)\|_2^2}{2\sigma_{\mathrm{rbf}}^2}\right),
\label{eq:rbf}
\end{equation}
so that similar trajectories receive larger kernel values. DPPs prefer subsets with large determinants, which favors sets that are both high-coverage and non-redundant in the embedding space~\cite{kulesza2012dpp}. We therefore select a subset $S^{(k)}$ of size $m$ by greedy approximation to
\begin{equation}
S^{(k)} \approx \arg\max_{|S|=m} \log\det\!\left(L_S+\varepsilon I\right),
\label{eq:dpp}
\end{equation}
where $L_S$ is the principal submatrix indexed by $S$ and $\varepsilon>0$ is a small regularizer. This yields a compact subset of trajectories that covers distinct recovery modes while avoiding near-duplicate rollouts~\cite{kulesza2012dpp,han2017faster}.

\subsection{Implementation Details}
In all experiments, distances are computed in a normalized task subspace, where position and orientation components are scaled to comparable numeric ranges before evaluating the Euclidean metric. Unless otherwise stated, we set $q_{\max}=0.8$ and choose $q_{\min}\in[0.1,0.3]$ based on task difficulty. If the number of successful rollouts in iteration $k$ is smaller than 5, we reuse the previous tube parameters $(r_{\min}^{(k-1)}, r_{\max}^{(k-1)})$ to avoid unstable quantile estimates. The exact parameters we used are shown in Table.~\ref{tab:pgdg_policy_params}.

\begin{table}[t]
\centering
\caption{PGDG Parameters}
\label{tab:pgdg_policy_params}
\small
\setlength{\tabcolsep}{4pt}
\begin{tabular}{p{0.38\columnwidth} p{0.52\columnwidth}}
\hline
\textbf{Parameter} & \textbf{Value / setting} \\
\hline
Inner quantile for $r_{\min}^{(k)}$ & $q_{\min}\in[0.1,0.3]$ \\
Outer quantile for $r_{\max}^{(k)}$ & $q_{\max}=0.8$ \\
Task subspace & $\psi(s)= [T_{\mathrm{obj}},\, T_{\mathrm{ee}}^{L},\, T_{\mathrm{ee}}^{R}]$ \\
Distance metric & Euclidean distance in normalized task subspace \\
Peak deviation & $d_{\mathrm{peak}}(\tau)=\max_t d_t(\tau)$ \\
Successful-rollout set & $\{\tau_n : \mathrm{Succ}(\tau_n)=1\}$ \\
Success-poor fallback & Reuse previous $(r_{\min}^{(k-1)}, r_{\max}^{(k-1)})$ if fewer than 5 successes \\
Max relabeled states & $K_{\mathrm{rel}}=10$ \\
Relabeling horizon & $H=15$ \\
Sampler stability term & $10^{-3}$ \\
Variance floor & $10^{-3}$ \\
\hline
\end{tabular}
\end{table}

\subsection{Pseudo Algorithm}

\begin{algorithm}[t]
\caption{PGDG: Physically Grounded Data Generation from One Demonstration}
\label{alg:pgdg}
\KwIn{expert demonstration $\tau_E$, rollout simulator $\mathrm{Rollout}(\cdot)$, outer iterations $K$, samples per iteration $N$}
\KwOut{training dataset $\mathcal{D}$}

Initialize proposal distribution $q^{(0)}(c)$ around $\tau_E$\;
Initialize curated trajectory set $\mathcal{B}_{\mathrm{sel}} \leftarrow \emptyset$\;
Initialize relabeled chunk set $\mathcal{Y}_{\mathrm{rel}} \leftarrow \emptyset$\;

\For{$k=0,\dots,K-1$}{
    Initialize successful batch $B^{(k)} \leftarrow \emptyset$\;

    \For{$n=1,\dots,N$}{
        Sample control parameter $c_n \sim q^{(k)}(c)$\;
        Decode nominal plan $a^{\mathrm{nom}}_{0:T-1} = G(c_n)$\;
        Roll out trajectory $\tau_n \leftarrow \mathrm{Rollout}(a^{\mathrm{nom}}_{0:T-1})$\;
        \If{$\mathrm{Succ}(\tau_n)=1$}{
            Add $(\tau_n,c_n)$ to $B^{(k)}$\;
        }
    }

    Compute adaptive tube bounds $r_{\min}^{(k)}, r_{\max}^{(k)}$ from successful peak deviations in $B^{(k)}$\;

    \ForEach{$(\tau_i,c_i)\in B^{(k)}$}{
        Compute curation reward $R^{(k)}(\tau_i)$\;
        Compute trajectory embedding $\phi(\tau_i)$\;
    }

    Select a diverse subset $S^{(k)} \subseteq B^{(k)}$ by greedy DPP using $\phi(\tau)$\;
    Update $q^{(k+1)}(c)$ from $\{c_i : (\tau_i,c_i)\in S^{(k)}\}$ by weighted moment matching with weights monotone in $R^{(k)}(\tau_i)$\;
    Add $\{\tau : (\tau,c)\in S^{(k)}\}$ to $\mathcal{B}_{\mathrm{sel}}$\;
}

Select the top-$K_{\mathrm{rel}}$ risky states across all trajectories in $\mathcal{B}_{\mathrm{sel}}$, using $d_t(\tau)$ with minimum temporal separation\;

\ForEach{selected relabel point $(\tau,t)$}{
    Extract reference segment $\bar u_{0:H-1}$ from $\tau$ starting at $t$\;
    Run short-horizon CEM to optimize $u^\star_{0:H-1}$ under $J(u;s_t)$ with full-episode success check\;
    Add $(o_t, y_t)$ to $\mathcal{Y}_{\mathrm{rel}}$, where $y_t=(u_0^\star,\dots,u_{H-1}^\star)$\;
}

Construct $\mathcal{D}$ from timestep-level pairs extracted from trajectories in $\mathcal{B}_{\mathrm{sel}}$, augmented with relabeled corrective targets in $\mathcal{Y}_{\mathrm{rel}}$\;
\Return{$\mathcal{D}$}
\end{algorithm}

\end{document}